\documentclass[letterpaper, 10 pt, conference]{ieeeconf}  

\IEEEoverridecommandlockouts                              

\usepackage{amsmath} 
\usepackage{svg}
\usepackage{pgfplots}
\pgfplotsset{compat=1.9}
\usepackage{graphicx}
\usepackage{caption}
\usepackage{xcolor}
\usepackage{textcomp}
\usepackage{caption}
\usepackage{subcaption}
\usepackage[
backend=biber,
style=ieee,
]{biblatex}

\title{\LARGE \bf
World Models for Anomaly Detection during Model-Based\\Reinforcement Learning Inference
}

\author{Fabian Domberg$^{1}$ and Georg Schildbach$^{1}$
\thanks{$^{1}$Autonomous Systems Lab (ASL), Institute for Electrical Engineering in Medicine,
        University of Lübeck, 23562 Lübeck, Germany
        {\tt\small f.domberg@uni-luebeck.de}}%
}

\begin{document}

\maketitle
\thispagestyle{empty}
\pagestyle{empty}

\begin{abstract}
Learning-based controllers are often purposefully kept out of real-world applications due to concerns about their safety and reliability. We explore how state-of-the-art world models in  Model-Based Reinforcement Learning can be utilized beyond the training phase to ensure a deployed policy only operates within regions of the state-space it is sufficiently familiar with. This is achieved by continuously monitoring discrepancies between a world model's predictions and observed system behavior during inference. It allows for triggering appropriate measures, such as an emergency stop, once an error threshold is surpassed. This does not require any task-specific knowledge and is thus universally applicable. Simulated experiments on established robot control tasks show the effectiveness of this method, recognizing changes in local robot geometry and global gravitational magnitude. Real-world experiments using an agile quadcopter further demonstrate the benefits of this approach by detecting unexpected forces acting on the vehicle. These results indicate how even in new and adverse conditions, safe and reliable operation of otherwise unpredictable learning-based controllers can be achieved.

\end{abstract}

\section{INTRODUCTION}
\begin{figure}[t!]
    \centering
    \hspace{0.3cm}\includegraphics[width=0.8\linewidth]{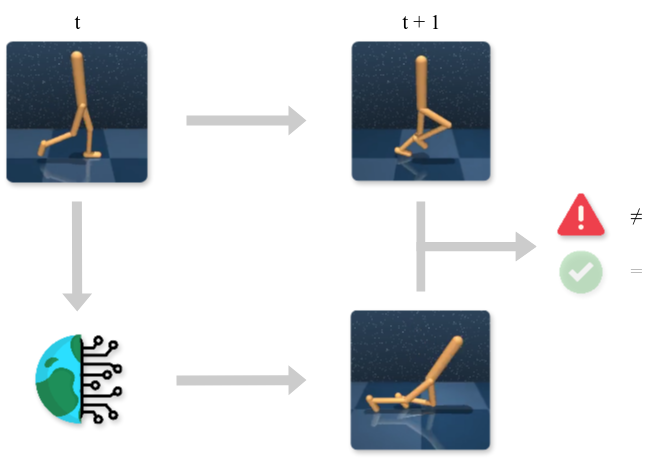}
    \caption{Illustration of our proposed method. Instead of discarding a world model after training, it can be used during inference to continuously generate predictions into the future. These can then be compared to the real outcomes and thus provide a measure about the familiarity of the agent with the situation at hand. This, in turn can be used to implement universal safety checks for learning-based controllers.}
    \label{fig:illustr}
\end{figure}

While artificially intelligent robots are achieving ever increasingly impressive results, their capabilities are still far from those of most animals and humans. Much like their disembodied relatives, such as chatbots and recommender systems, today's robotic systems, be it industrial robot arms or dog-like inspection vehicles, still fall well within the category of \emph{narrow} intelligence. That is, they can only perform a very specific set of tasks. While they are often superhumanly capable at one, they are entirely incapable of others. This can largely be attributed to a lack of generalization which biological systems, especially humans, excel at. We argue that a first step toward solving this issue is to improve their situational awareness such that they are able to tell whether a given situation is different from what they are used to. Our proposed method is illustrated in Figure \ref{fig:illustr}.

\section{RELATED WORK}
Although there is ongoing research on how to broaden the capabilities of intelligent robots, for example to be able to carry out multiple previously unknown tasks \cite{team2021open}, some argue there is something fundamental missing in today's approaches.
For example, in their $2022$ work \emph{A Path Towards Autonomous Machine Intelligence} Y. LeCun argue that the levels of intelligence seen in advanced biological systems may only be attainable through the interplay of multiple different modules in the brain \cite{lecun2022path}. These include some that are already present in today's robot controllers, such as a perception module that is responsible for pre-processing inputs and an actor module which takes these in to produce decisions on what to do next. This basic control loop is oftentimes sufficient for simple tasks. A more complicated task, however, may require a memory module, for example to execute consecutive tasks in a specific order. In current state-of-the-art systems this is usually implemented using a Recurrent Neural Network, such as an LSTM. Even though these existing modules allow for impressive behaviors, the overall systems using them today still lack a fundamental capability, LeCun reasons. That is, using a perception module allows to make decisions based only on the present and a memory module enables to consider the past. Biological intelligence, however, also considers the future. By planning, i.e. mentally exploring and evaluating future system states and outcomes, it can solve significantly more complex tasks. Though this requires what LeCun describes as a ``learnable internal model of how the world works''. Such a \emph{world model} module would have the ability to recursively predict the next state given a prior one. Ideally, this would result in realistic sequences of entirely predicted state-action pairs, allowing to explore different routes through the state-space without actually acting them out. A system with this ability could refine its decisions by simulating multiple potential scenarios, selecting the most promising course of action. This is similar to \emph{System 2} thinking in humans \cite{kahneman2011thinking}.

\subsection{World Models in Reinforcement Learning}
Though usually not yet referred to as a world model, the idea of learning a predictive model that can be interacted with in place of the real environment has been studied within the field of Model-Based Reinforcement Learning (MBRL) for decades \cite{arulkumaran2017brief}. Although this has seen some success, mostly in the form of increasing training sample efficiency, results were often limited to simple tasks with low dimensional action and observation spaces. However, in $2018$ Ha \& Schmidhuber were able to leverage recent advances in deep learning and computer vision to train an auto-encoder-based world model capable of predicting future image outputs of a videogame \cite{ha2018recurrent}. The actual predicting thereby takes place in the low dimensional latent space of the auto-encoder, attached to a memory unit to capture temporal relations. Due to this, they are able to optimize their policy much more efficiently by entirely training within the world model, instead of the relatively resource hungry videogame.


Building upon this proof-of-concept, Hafner et al. proposed the \emph{Dreamer} MBRL algorithm in $2019$ \cite{hafner2019dream}. Through the combination of multiple state-of-the-art techniques, Dreamer handles control tasks, especially those with visual input, much more effectively than previous approaches. In addition to predicting future states, its world model also considers future state values and rewards. With DreamerV3, the most recent iteration, they show favorable scaling behavior for model parameter count and the ratio between world model and real environment interactions \cite{hafner2023mastering}. They show how these findings enable Dreamer to thrive especially in complex, high-dimensional state-space tasks with sparse rewards, such as $3$D videogames.

\subsection{World Models for Anomaly Detection}
The concept of world models has been found to be useful in fields other than MBRL as well. A notable example is video generation. Here, it is of upmost importance that objects shown behave according to the laws of physics. Garrido et al. recently showed how a model fulfilling this can be trained in an unsupervised way \cite{garrido2025intuitive}. They leverage the violation-of-expectation framework from developmental psychology to identify model errors without requiring any task-specific knowledge. This models how human infants learn through observing their surroundings and updating their mental model if anything unexpected occurs. In $2006$, Silva et al. first proposed the use of this kind of prediction error to detect model mismatch in a non-stationary MBRL setting \cite{da2006dealing}. That is, they use a confidence measure, derived from state and reward errors, to recognize changes in the agent's environment and react accordingly. More recently, Bogdoll et al. presented a conceptual study on how modern MBRL algorithms, including DreamerV3, may be used for anomaly detection in an autonomous driving context \cite{bogdoll2023exploring}.

\subsection{Contribution}
This work examines the use of world models during policy inference. Specifically, examining whether current world models provide the necessary fidelity to detect modifications in the learning environment. It is explored if and how this could be used to equip autonomous agents with a built-in safety mechanism, allowing them to gauge their familiarity with a situation and whether or not they can successfully act in it. The results are verified in both simulated and real-world experiments.

\section{METHODOLOGY}
The following section presents our approach in detail, including the underlying theoretical considerations and assumptions, as well as describing the experimental setups.

\subsection{Rationale}
In classic MBRL world models are used solely during the training phase. They serve as synthetic data generators for an agent to interact with instead of the real environment, significantly improving sample efficiency and thus enabling it to solve more complex problems with less time and resources \cite{hafner2023mastering}. After training, however, they are usually discarded. We propose to also use a trained agent's world model during inference, allowing it to make predictions for how the system state is going to evolve over the next few timesteps. Once these timesteps have passed in reality, the prediction error can be computed. As shown in previous works, this can then be used to infer the presence of anomalies, i.e., deviations from the expected outcome. This may, for example, be changes in the environment such as a slippery surface a robot moves onto, or changes to the robot itself such as a defect in one of its actuators. Detecting such events is crucial for systems and controllers that work outside of controlled environments. This holds especially for \emph{black box} controllers, like neural-networks, because such systems are notorious for behaving unpredictably once they find themselves in unfamiliar states. We therefore argue that this prediction error may be understood as an indirect reliability and safety measure for an agent's outputs and decisions.

\subsection{Theoretical Considerations and Assumptions}
\label{subsec:theory}
Since DreamerV3's world model is essentially an auto-encoder, it is very effective at encoding and decoding the data it is exposed to during training. When given data outside the original distribution, however, reconstruction performance quickly deteriorates. This poor generalization is inherent to auto-encoders, as their latent space is purposefully kept as small as possible, i.e., to the minimum required to adequately reconstruct the training data. While this characteristic is undesirable for most use-cases, it can serve as a measure for ‘distance’ from the current input data to the training data distribution. In other words, a significantly higher reconstruction error likely means that the current input data is not ‘close’ to those observed during training. Since DreamerV3's world model and policy are trained simultaneously, and are thus exposed to the same training data, it will therefore be assumed that any noticeable prediction error by the world model during inference coincides with an observation the policy is not sufficiently familiar with in order to generate a productive decision.


\subsection{Implementation}
\label{sect:impl}
During inference of a fully trained DreamerV3 agent, predictions into the future are obtained in the same way as during the training phase \cite{hafner2023mastering}. That is, at every timestep $t$, the encoder compresses raw observations $x_t$ into a stochastic representation $z_t$. After feeding $z_t$ through the recurrent sequence model, the recurrent state $h_t$ is obtained. Together, $z_t$ and $h_t$ make up the internal model state $s_t$. Given this, an action $a_t$ is sampled from the actor policy $\pi_\theta(a_t, s_t)$. Using $s_t$ and $a_t$, the world model can then be used to predict the next state $\hat{s}_{t+1}$, which in turn can be used to obtain $\hat{a}_{t+1}$ and any further $\hat{s}_{t+n}$, recursively. A prediction horizon of $n=16$ is used. For each imagined step $i \in [1, n]$ of the prediction, $\hat{s}_{t+i}$ is fed through the decoder to reconstruct the corresponding predicted observation $\hat{x}_{t+i}$, as well as the predicted reward $\hat{r}_{t+i}$. Later, after $n$ real timesteps have passed and the true observations and rewards are collected, the observation and reward error for a single timestep $t$ are computed as the average absolute difference over the entire prediction horizon:
\begin{equation}
\label{eq:pred_err}
    e_{\text{obs}_t} = \frac{1}{n} \sum_{i=1}^{n} | \hat{x}_{t+i} - x_{t+i} | ,
\end{equation}
\begin{equation}
\label{eq:rew_err}
    e_{{rew}_t} = \frac{1}{n} \sum_{i=1}^{n} | \hat{r}_{t+i} - r_{t+i} |.
\end{equation}

For image-based state-spaces, the error between a predicted $\hat{I}_t$ and observed $I_t$ image is itself an image:

\begin{equation}
    E_{\text{img}_t} = | \hat{I}_t - I_t |.
\end{equation}
This is mainly for illustration purposes, as it can be interpreted as an error heatmap and thus also provide positional information. To obtain a singular value, one could compute the average individual pixel error as above. Though it would be advised to utilize more suitable image comparison metrics instead, since raw pixel-wise differences may not align with perceptual quality.


\section{RESULTS}
In each of the following experiments, the learning algorithm is SheepRL's implementation of DreamerV3 \cite{EclecticSheep_and_Angioni_SheepRL_2023}. We use the \emph{small} model preset as these environments are relatively simple. If not stated otherwise, default parameters are used. Each training is run until convergence. After training, we run inference and simultaneously compute the world models' prediction errors as described in \ref{sect:impl}.

\begin{figure}[hb!]
    \centering
    \hspace*{-0.24cm}\hspace*{0.32cm}\includegraphics[width=.89\linewidth]{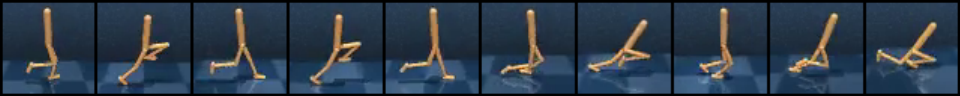}
    \resizebox{1.\linewidth}{!}{\input{"figs/walker2.pgf"}}
    \caption{DMC's Walker-walk task with DreamerV3's world model's predictions (green) and actual measurements (blue), along with corresponding smoothed observation and reward error. After $500$ timesteps, gravity is increased by $50\%$.}
    \label{fig:exp_walker}
\end{figure}

\subsection{Simulated Experiments}
To illustrate the details of our proposed method, we demonstrate it on two exemplary tasks from the well established Deep Mind Control (DMC) suite \cite{tassa2018deepmind}. We train a regular DreamerV3 agent to perform these and, after training is complete, observe its world model's predictions while abruptly changing parts of the simulation. Each example represents one of the two general categories all of our conducted experiments fall into: \emph{global} and \emph{local} modifications. Figure \ref{fig:exp_walker} depicts an experiment in the former category. It shows chronological screenshots of the task being performed at the top, in this case \emph{Walker-walk}, and a selection of corresponding state variables' behavior over time below. For the first $500$ timesteps, the environment is kept in its default configuration, i.e. the same as during training. During this time, predictions and actual measurements closely match each other, as expected from a fully trained world model. There is a smoothed average observation error below or equal to two and no notable reward error. The human-like Walker figurine walks to the right as it was trained to do, raising its legs rhythmically and putting one foot in front of the other to propel itself forward. After the $500$ timesteps have passed, gravity is increased by $50 \%$. Thereafter, the Walker quickly drops to its knees from the new forces acting on it. Throughout the remaining timesteps it repeatedly tries to raise its body upright, only to keep tripping back down. Each fall coincides with an increase in the observation error. Some individual state variables now also show a deviation between prediction and reality, for example, the torso's and left leg's $z$ position. Others, such as the right foot's $x$ position, show no notable change. Additionally, there is a clear increase in the smoothed reward error as the Walker is unable to maintain its expected posture and velocity. Even without knowing the root cause, this information clearly indicates an impediment for the agent to continue its task safely.

\begin{figure}[ht!]
    \centering
    \hspace*{-0.24cm}\hspace*{0.32cm}\includegraphics[width=.9\linewidth]{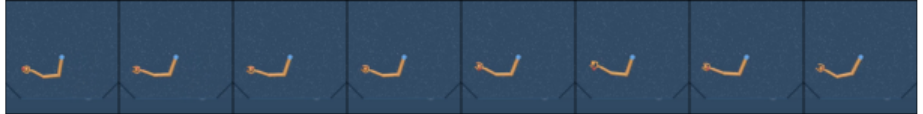}
    \resizebox{1.\linewidth}{!}{\input{"figs/bringball2.pgf"}}
    \caption{DMC's Manipulator-bring-ball task with DreamerV3's world model's predictions (green) and actual measurements (blue), along with corresponding smoothed observation and reward error. At timestep $480$ the gear ratio of the elbow joint is increased by a factor of three.}
    \label{fig:exp_bringball}
\end{figure}

Figure \ref{fig:exp_bringball} shows results from an experiment in the category of local modifications. The underlying DMC task is called \emph{Manipulator-bring-ball}, where the goal of a robot arm is to pick up and lift a ball to a target location and keep it there.\\
Again, during the first $480$ timesteps using the default configuration, the agent fulfills the task as intended. During the first $100$ timesteps, the arm finds the ball, picks it up and brings it to the desired location. Corresponding deflections are seen in the respective state variables. There is also a peak of around $1.5$ in the smoothed observation error. Once the ball reaches its target, most state variables' predictions closely match the actual measurements and the smoothed observation error drops to around $0.4$. At timestep $480$, the gear ratio of the elbow joint, i.e., between first and second arm element, is increased by a factor of three. This has the effect that the joint rotates more than it used to, given the same control input. This intends to simulate a sudden actuator defect. It shows the most prominent prediction discrepancy exactly where it is applied: in the rotational velocity of the elbow. Up- and downstream arm elements, such as the root's $x$ position or the wrist's rotational velocity, on the other hand show little to no change. The only other state variables that hint at the sudden oscillatory movements of the lower arm around the target location are those related to the ball. That is because rather than gently being guided by the arm's two-fingered hand, it violently bounces around in it. While somewhat noticeable in the ball's $x$ velocity, this is especially evident from the fingertip sensors, which are suddenly seeing unexpected high frequency activations. The smoothed observation error more than doubles to circa $1.0$ and there is a stark increase in the reward error.

\begin{figure}[ht!]
    \centering
    \subfloat[Averaged Inputs]{{\includegraphics[width=.99\linewidth]{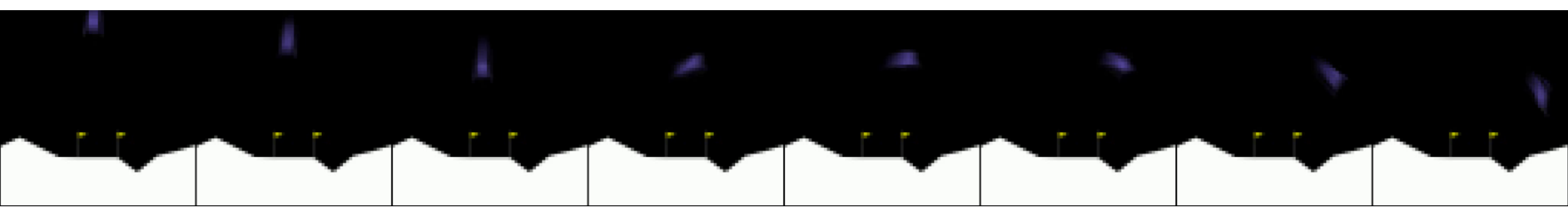}}}
    \qquad
    \subfloat[Averaged Predictions]{{\includegraphics[width=.99\linewidth]{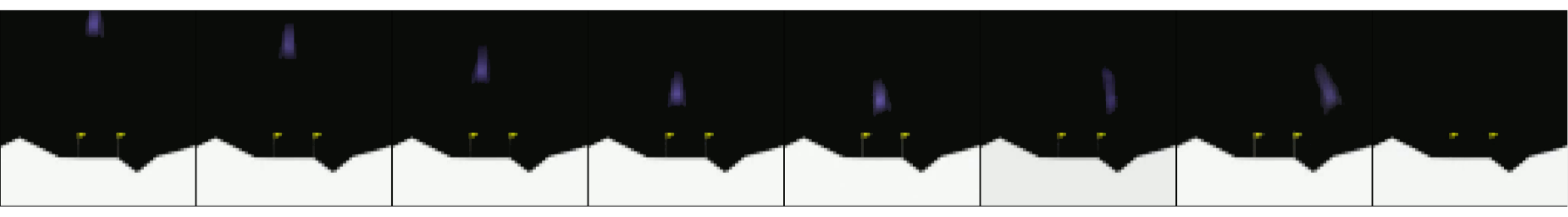} }}
    \qquad
    \subfloat[Errors]{{\includegraphics[width=.99\linewidth]{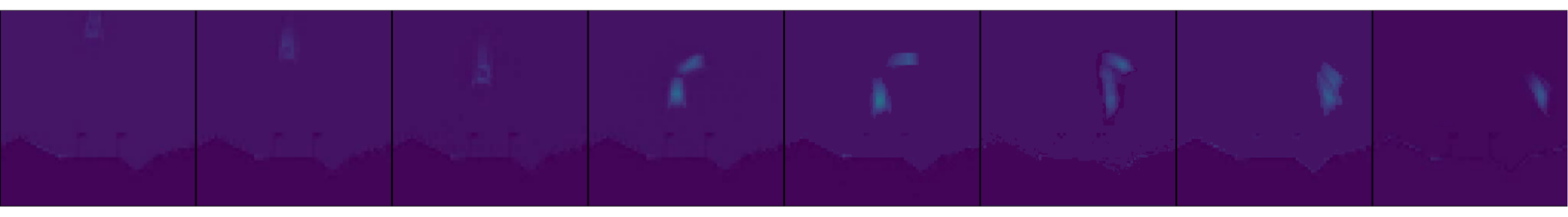}}}

    \caption{DreamerV3's world model with visual input reacting to an unexpected force applied to the spacecraft in \emph{LunarLander-v2} after the third image, pushing it to the top-right. Both input and prediction images are averaged over $16$ timesteps to visualize movement. The resulting error heatmap displays the location of unexpected events and their magnitude. For visualization purposes, as well as noise reduction, we render multiple images into one here.} 
    \label{fig:exp_visual}
\end{figure}

\subsection{Visual Observations}
\label{sect:visual}
We demonstrate that our method can even be used with visual observation spaces. Figure \ref{fig:exp_visual} depicts an experiment using the \emph{LunarLander-v2} simulation \cite{towers2024gymnasium}. Again, a standard DreamerV3 agent is trained to solve the task - in this case using the blue spacecrafts' left, right and down thrusters to gently land on the lunar surface within the two yellow markers. To do so, it is given only the current image of the scene. Each of the square images in Figure \ref{fig:exp_visual} shows the spacecrafts position averaged over the next $n=16$ timesteps. In case of the predictions, it shows how the world model thinks the simulation is going to evolve, given observation images only up to this point. The error image represents the absolute difference between the two as a heatmap. After the third image, a force is applied to the spacecraft, pushing it to the top right. This is where a clear discrepancy between reality and prediction appears for the first time. While in the input images the spacecraft can be seen flying towards the right, the predictions show it continuing its decent. Over the next timesteps they slowly converge back together. In the final image, the spacecraft has moved so far to the right that the world model can no longer model it, which in this case leads to its predicted disappearance. This position of the spacecraft represents an out-of-distribution sample, i.e., one that has not sufficiently been seen during training. These are precisely what makes operating learning-based controllers dangerous to operate outside of controlled laboratory environments.

\begin{figure*}[ht!]
    \centering
    \begin{subfigure}[t]{0.49\linewidth}
        \centering
        \hspace*{0.09cm}\includegraphics[width=.845\linewidth]{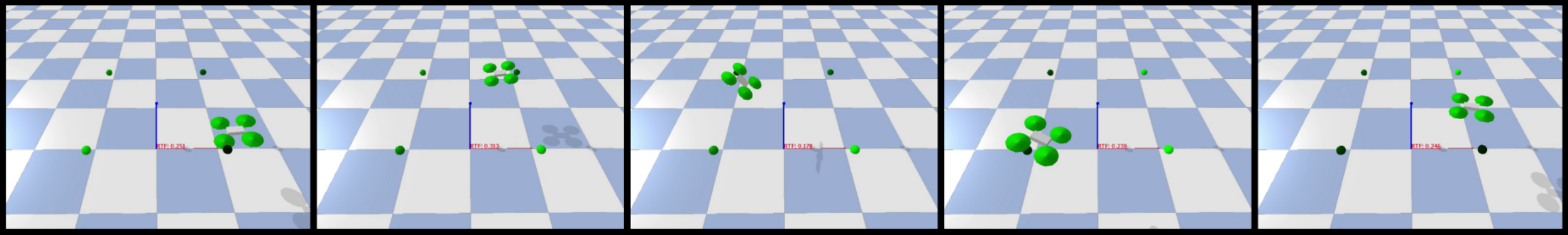}
        \resizebox{.95\linewidth}{!}{\input{"figs/simSq2.pgf"}}
        \caption{Simulation}
        \label{fig:exp_sim}
    \end{subfigure}
    \hfill
    \begin{subfigure}[t]{0.49\linewidth}
        \centering
         \hspace*{0.09cm}\includegraphics[width=.86\linewidth]{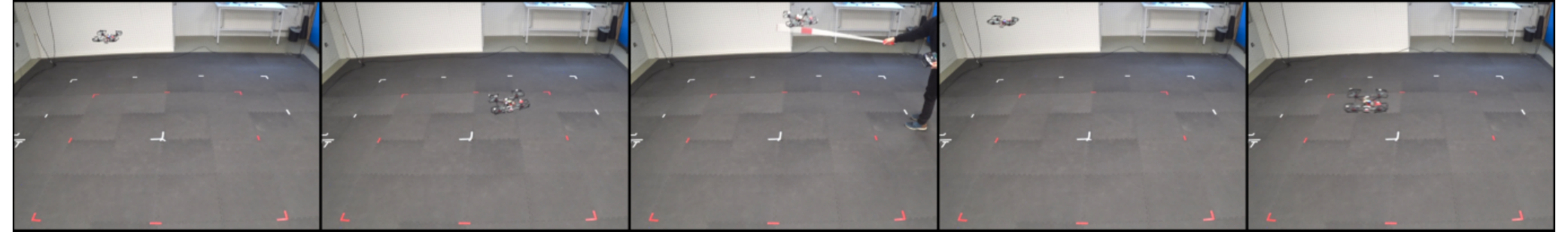}
        \resizebox{.95\linewidth}{!}{\input{"figs/labSq2.pgf"}}
        \caption{Real-world}
        \label{fig:exp_lab}
    \end{subfigure}
    \caption{The same experiment performed in simulation and the real-world. A quadcopter successively navigates the corners of a $1$ m\textsuperscript{2} square. At $240$ and $205$ timesteps, respectively, a sudden force briefly acts upon it. While world model predictions (color) and true measurements (color) closely align throughout the rest of the flight, this event produces a significant error.}
    \label{fig:exp_combined}
\end{figure*}

\subsection{Real-World Validation}
To demonstrate the practicality for real-world robotics, we conduct the following experiment twice: once in simulation and once in reality. A $600$ g quadcopter is tasked with sequentially navigating the corners of a $1$ m\textsuperscript{2} square. After some time, an abrupt force acts upon it. In the real-world experiment, this is implemented by hitting the drone with a broomstick. The agent's observation are made up of the quadcopter position, velocity, acceleration and angular velocity for each of the three axes, respectively, as well as the last action and the two next waypoints coordinates. Its four outputs control roll, pitch, yaw and throttle. The last action and next two waypoints are omitted from the following graphs and are also not included in the observation error.
Figure \ref{fig:exp_sim} shows the experiment conducted in the \emph{PyFlyt} simulation \cite{tai2023pyflyt}. During the initial $240$ timesteps, the world model's predictions closely align with the measured values and the observation error hovers below $0.25$. Then, as the unexpected force is applied, there is a notable divergence in most state variables' predictions, resulting in a peak of over $0.5$ in observation error. After rebalancing, error levels return to normal.
A similar but noticeably different behavior can be observed in Figure \ref{fig:exp_lab}, which portraits the real-world experiment. Here, predictions and measurements do not align as closely as in the simulation. This is because the simulation can only model the system with finite accuracy. The observation error exhibits more fluctuations with increased magnitude. Though exactly as in simulation, there is a clear peak in observation error once the drone is nudged with the stick at around timestep $205$. As before, after rebalancing, the observation error returns to its prior level. The only state variable with a permanent mismatch in prediction and measurement is the velocity in $z$ direction, showing an obvious delay likely due to the drone having a lower climb rate than its simulated counterpart.
Regarding reward error, the simulated experiment only shows a small increase upon the force's impact, while in the real-world experiment there is a slight decrease. These are due to the way the reward function in this particular case is defined, giving small rewards for continuous waypoint approach and large ones for waypoint traversal. Depending on the drone's distance to a waypoint when it is struck, this can lead to different prediction results. This emphasizes the fact that our method can only detect abnormal situations, not interpret their meaning. For this, task-specific knowledge must additionally be considered.



\section{DISCUSSION}
\subsection{Overall Error may be too Vague}
While most changes to the environments during our experiments could be detected successfully as an increase in the average observation error, there were some that could not be distinguished from noise. Oftentimes, however, there did indeed occur noticeable changes to at least some individual state variable. Specifically, global changes often lead to small but widespread prediction errors across many state variables. In contrast, local modifications tend to produce more focused and isolated prediction discrepancies. They primarily affect state variables directly connected to the modification, with downstream effects limited to components closely tied to the altered element. Monitoring individual state variables or cluster thereof may thus be better suited in these cases.

\subsection{Prediction Errors must be Interpreted}
Ultimately, a prediction error alone merely indicates deviations from an expected outcome, it does not provide any insight into the underlying sources of this discrepancy, nor does it assess its relevance, impact, or practical significance. This means a prediction error is not to be interpreted as a direct measure of safety or reliability, but rather requires another layer of interpretation. This may be realized through the use of prior knowledge, for example by weighting prediction errors of certain state-space variables differently. For the example of a car, this could mean paying less attention to an error in longitudinal direction, as this can be relatively easily corrected, but more toward lateral deviations as this degree of freedom cannot be controlled directly and thus poses a higher threat to overall system safety.
If there is no task-specific knowledge available, one approach could be to feed the prediction errors back to the agent and train it to react accordingly. This solution is more general and applicable to more complex systems. It may be explored in future work.

\subsection{Applicable to Visual Observation Spaces}
In Section \ref{sect:visual} we show how our method can also be used with images. The experiment conducted shows how the resulting error image can be interpreted as a heatmap, indicating location and magnitude of the unexpected event. Though, again, the meaning of this can only be judged by considering task-specific knowledge. Also, future work may utilize more sophisticated image similarity metrics to obtain even richer insights. Additionally, we hypothesize that a combination of visual and numeric observation spaces could further improve world model fidelity.

\subsection{Benefits for Sim2Real Transfer and Interpretability}
During transfer of the agent from simulation to reality, the presented approach facilitated the process because implementation issues, such as flipped coordinate frames or mismatched control outputs, were rapidly identified from prediction errors. Higher level issues, such as oscillations introduced by the drone's internal PID controller or latency in sensor readings, were also easily diagnosed and corrected. An example of this is the delay in $z$ velocity in Figure \ref{fig:exp_lab}, originating from a mismatch between the real and simulated quadcopters' climb-rate. Identifying problems in this way not only allows for quick debugging, but also aids with understanding why an agent would misbehave in certain situations. This fact could be used to improve neural controller interpretability in future research.

\section{CONCLUSION}
 Our experiments show that both local and global changes, such as individual actuator failure and gravitational magnitude, can be detected as a discrepancy between world model prediction and measured reality. We caution that the mere existence of such model error does not necessarily imply any dangerous situation or agent misbehavior. Though it could aid future artificially intelligent agents to recognize and investigate inaccuracies in their understanding of the world around them. This, in turn, may eventually enable them to self-correct and improve. These hypotheses align with recent findings in frontier artificial intelligence research, in which an increase in test-time compute is shown to coincide with enhanced agent capability and even self-improvement \cite{guo2025deepseek}.\\

\addtolength{\textheight}{-12cm}   






\printbibliography 
\end{document}